\documentclass[letterpaper, 10 pt, conference]{ieeeconf}  

\IEEEoverridecommandlockouts                              

\title{\LARGE \bf
EPN: An Ego Vehicle Planning-Informed Network for Target Trajectory Prediction
}

\author{Saiqian Peng$^{1}$, Duanfeng Chu$^{1}$,~\IEEEmembership{Member,~IEEE,}, Guanjie Li$^{2}$, Liping Lu$^{2}$ and Jinxiang Wang$^{3}$,~\IEEEmembership{Member,~IEEE}
\thanks{*This work was supported in part by the National Key R\&D Program of China (2022YFB2502904), the Natural Science 
Foundation of Hubei Province for Distinguished Young Scholars (2022CFA091), the Key R\&D Program of Hubei 
Province (2024BAB033), and Wuhan Science and Technology Major Project (2022013702025184). \it{(Corresponding author: Duanfeng Chu.)}}
\thanks{$^{1}$Saiqian Peng and Duanfeng Chu are with the Intelligent Transportation Systems Research Center, Wuhan University of Technology, Wuhan 430063, China
        {\tt\small psq@whut.edu.cn, chudf@whut.edu.cn}}%
\thanks{$^{2}$Guanjie Li and Liping Lu are with the School of Computer Science and Artificial Intelligence, Wuhan University of Technology, Wuhan 430070, China
        {\tt\small guanjieli@whut.edu.cn, luliping@whut.edu.cn}}%
\thanks{$^{3}$Jinxiang Wang is with the School of Mechnical Enginerring, Southeast University, Nanjing 211189, China
        {\tt\small wangjx@seu.edu.cn}}%
}

\usepackage{url}
\usepackage{graphicx}
\usepackage{booktabs}
\usepackage{array}
\usepackage{multirow}

\begin{document}

\maketitle
\thispagestyle{empty}
\pagestyle{empty}

\begin{abstract}

Trajectory prediction plays a crucial role in improving the safety of autonomous vehicles. However, due to the highly dynamic and multimodal nature of the task, accurately predicting the future trajectory of a target vehicle remains a significant challenge. To address this challenge, we propose an Ego vehicle Planning-informed Network (EPN) for multimodal trajectory prediction. In real-world driving, the future trajectory of a vehicle is influenced not only by its own historical trajectory, but also by the behavior of other vehicles. So, we incorporate the future planned trajectory of the ego vehicle as an additional input to simulate the mutual influence between vehicles. Furthermore, to tackle the challenges of intention ambiguity and large prediction errors often encountered in methods based on driving intentions, we propose an endpoint prediction module for the target vehicle. This module predicts the target vehicle endpoints, refines them using a correction mechanism, and generates a multimodal predicted trajectory. Experimental results demonstrate that EPN achieves an average reduction of 34.9\%, 30.7\%, and 30.4\% in RMSE, ADE, and FDE on the NGSIM dataset, and an average reduction of 64.6\%, 64.5\%, and 64.3\% in RMSE, ADE, and FDE on the HighD dataset. The code will be open sourced after the letter is accepted.

\end{abstract}

\section{INTRODUCTION}

With the rapid development of autonomous driving technology, the safety of autonomous vehicles has garnered increasing attention. Accurately predicting the future trajectory of surrounding vehicles is crucial for the safe operation of autonomous vehicles \cite{Huang2022sur}. As an upstream module of planning, trajectory prediction plays a key role in supporting planning tasks and is one of the critical components in enhancing the safety of autonomous driving systems. However, due to the high dynamic and multimodal nature of trajectory prediction tasks, accurately forecasting the future trajectory of a target vehicle remains a significant challenge. High dynamism refers to the influence of other traffic participants on the target vehicle, which can cause the trajectory of the target vehicle to change unexpectedly. Multimodality refers to the possibility of multiple plausible future trajectories for a given historical trajectory. These characteristics make trajectory prediction a complex and difficult task.

To address these challenges, we propose an Ego vehicle Planning-informed Network (EPN) for multimodal trajectory prediction. In response to the high dynamic characteristics of trajectory prediction, current trajectory prediction methods typically use the historical trajectory and vehicle attributes as inputs, focusing primarily on how historical information influences the future trajectory of the target vehicle. However, in real-world driving scenarios, the future trajectory of a vehicle is influenced not only by its own historical data, but also by the behavior of other vehicles on the road. So, we incorporate the future planned trajectory of the ego vehicle as an additional input to model the mutual influence between vehicles. Furthermore, to account for the multimodal nature of trajectory prediction, we introduce a target endpoint prediction module that predicts multiple plausible target endpoints, thereby generating more realistic and accurate trajectory predictions for the target vehicle. Our contributions can be summarized as follows:

\begin{enumerate}
    \item We propose a multimodal trajectory prediction model based on ego vehicle planning, significantly improving the prediction performance of mapless trajectory prediction methods.
    \item We introduce a feature fusion encoding module that first extracts temporal features from the input information using a Long Short-Term Memory (LSTM) encoder, then models interactions between various traffic participants with a convolutional social pooling network to capture spatial features. The module outputs a unified feature vector that integrates both temporal and spatial characteristics.
    \item We propose a target's endpoint prediction module that first predicts the possible endpoints of the target vehicle using a Conditional Variational Autoencoder (CVAE), then refines the preliminary predicted endpoints coordinates with a correction mechanism, and finally uses the corrected endpoints to predict the complete trajectory of the target vehicle.
\end{enumerate}

\section{RELATED WORK}
\label{section2}
\subsection{Physics-based methods}
In the early stages of trajectory prediction technology, single trajectory methods were commonly used. Dynamic models made short-term predictions by describing the vehicle's dynamic characteristics \cite{Lin2000dyn}, \cite{Kaempchen2009}. Kinematic models are mainly based on kinematic attributes such as vehicle speed, acceleration, and steering angle \cite{Schubert2008}, \cite{Lytrivis2008}. These methods primarily focus on the state of the target vehicle and fail to account for the influence of other factors, such as surrounding vehicles and environmental conditions. The Kalman filter method incorporates noise consideration within the model construction and predicts vehicle trajectories through Gaussian distribution modeling \cite{Jin2015}, \cite{Dyckmanns2011}. Compared to the Kalman filter, which is limited to linear scenarios, the Monte Carlo method offers advantages in handling complex nonlinear and multimodal scenarios \cite{Wang2019MPC}.

\subsection{Machine learning methods}
As technology has advanced, machine learning methods have been increasingly applied to vehicle trajectory prediction. Unlike physics-based models, machine learning methods are data-driven approaches. Notable methods include Gaussian Processes (GPs), Support Vector Machines (SVMs), Hidden Markov Models (HMMs), and Dynamic Bayesian Networks (DBNs). GPs use historical trajectory data to learn the potential distribution of future trajectories \cite{Guo2019GRF}. SVMs classify data by finding an optimal hyperplane, helping to determine driving intentions such as left turns, straight movements, and right turns \cite{Aoude2010}. HMMs effectively simulate temporal changes and uncertainties in driving behavior by modeling vehicle driving patterns as hidden states, but they are limited to systems with discrete states \cite{Qiao2015Ada}. DBNs are well-suited for handling complex multidimensional states and long-term dependencies, but they can suffer from significant errors when converting recognized intentions into accurate trajectories \cite{Li2019bay}.

\subsection{Deep learning methods}
Deep learning has revolutionized trajectory prediction by enabling comprehensive modeling of vehicle dynamics, social interactions, and environmental constraints through neural networks' parametric capacity \cite{Xu2023Context}, \cite{Deo2018Conv}, \cite{Gupta2018sgan}, \cite{Li2021GANTLSTF}, \cite{Xin2018Intention}, \cite{Vishnu2023Traffic}, \cite{Song2020pip}, \cite{Guo2022ego}, \cite{Sheng2024GCN}, \cite{Feng2023MacFormer}, \cite{Liu2023stf}, \cite{Schmidt2022CRAT}, \cite{Gu2021Densetnt}, \cite{Aydemir2023ADAPT}, \cite{Karim2024Destine}. Early works established interaction modeling frameworks using convolutional social pooling \cite{Deo2018Conv} and GANs \cite{Gupta2018sgan}, \cite{Li2021GANTLSTF}, while \cite{Xin2018Intention} pioneered intention-aware prediction through dual-LSTM architectures. Recent advances emphasize the critical role of additional input information. Song et al. \cite{Song2020pip} and Guo et al. \cite{Guo2022ego} demonstrated significant accuracy gains by encoding ego vehicle plans via LSTM and attention mechanisms, with Sheng et al. \cite{Sheng2024GCN} further reducing behavioral uncertainty through explicit trajectory encoding in graph networks. Another important aspect that influences performance is the output formulation of prediction models, where current multimodal approaches diverge into two distinct paradigms. Intention-based methods, such as those in \cite{Song2020pip}, \cite{Guo2022ego}, \cite{Liu2023stf}, and \cite{Schmidt2022CRAT}, classify discrete driving maneuvers to generate corresponding trajectories. In contrast, endpoint-driven approaches, such as those in \cite{Gu2021Densetnt}, \cite{Aydemir2023ADAPT}, and \cite{Karim2024Destine}, demonstrate superior performance through endpoint-constrained trajectory generation.

\section{PROBLEM FORMULATION}
\label{section3}
\begin{figure}[!t]
\centering
\includegraphics[height=2.5cm,width=7cm]{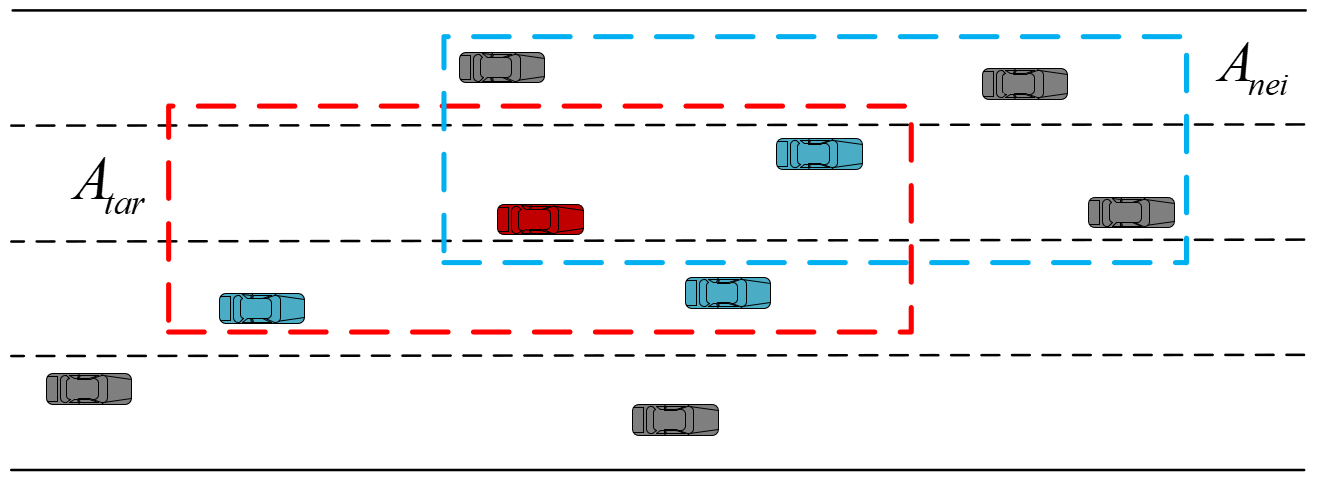}
 \caption{Vehicle classification and region of interest division in a driving scenario.}
 \label{fig1}
\end{figure}

As illustrated in Figure \ref{fig1}, vehicles in the driving scenario are classified into three types: the red vehicle represents the ego vehicle, blue represents the target vehicle, and gray represents adjacent vehicles. The area of interest is divided into a grid of 200 by 35 feet, centered on the ego vehicle, denoted as $A_{tar}$, which corresponds to the region within the red dashed line. Similarly, the area surrounding the target vehicle, referred to as $A_{ner}$, is shown within the blue dashed line in the figure.

\subsection{Model Input}
The model's input consists of the historical driving state information of all vehicles in the scenario, represented as:

\begin{equation}
    {X^i} = \{ s_{t - {T_h} + 1}^i,s_{t - {T_h} + 2}^i,...,s_t^i\}
\end{equation}
where $i$ denotes the vehicle index, $T_h$ represents the historical time steps used as input, and $s_t^i$ is the driving state of the $i$-th vehicle at time step $t$, defined as:
\begin{equation}
    s_t^i = (x_t^i,y_t^i,v_t^i,a_t^i)
\end{equation}
where $x_t^i,y_t^i,v_t^i,$and $a_t^i$ represent the horizontal position, vertical position, velocity, and acceleration of vehicle $i$ at time step $t$, respectively.

Additionally, the model input includes the future planned trajectory of the ego vehicle, given by:
\begin{equation}
    P = \{ {p_{t + 1}},{p_{t + 2}},...,{p_{t + {T_f}}}\}
\end{equation}
where $T_f$ denotes the future prediction horizon, and $p_{t+T_{f}}$  is the planned position of the ego vehicle at time step $(t+T_{f})$, defined as:
\begin{equation}
    {p_{t + T_{f}}} = ({x_{t + T_{f}}},{y_{t + T_{f}}})
\end{equation}
where $x_{t + T_{f}}$,$y_{t + T_{f}}$ are the horizontal and vertical coordinates of the ego vehicle at time $(t+T_{f})$, respectively.

\subsection{Model Output}
The model's output is the predicted position sequence of the target vehicle, represented as:
\begin{equation}
    \widehat Y = \{ \widehat f_{t + 1}^i,\widehat f_{t + 2}^i,...,\widehat f_{t + {T_f}}^i\} 
\end{equation}
where $\widehat f_{t + {T_f}}^i$ denotes the predicted trajectory of the $i$-th target vehicle at time step $(t+T_{f})$. This predicted trajectory can be further expressed as:
\begin{equation}
    \widehat f_{t + {T_f}}^i = (\widehat x_{t + {T_f}}^i,\widehat y_{t + {T_f}}^i)
\end{equation}
where $\widehat x_{t + {T_f}}^i$ and $\widehat y_{t + {T_f}}^i$ represent the predicted horizontal and vertical coordinates of the $i$-th target vehicle at time step $(t+T_{f})$, respectively.

\section{MODEL ARCHITECTURE}
\label{section4}
\begin{figure*}[!t]
\centering
\includegraphics[height=7cm,width=16cm]{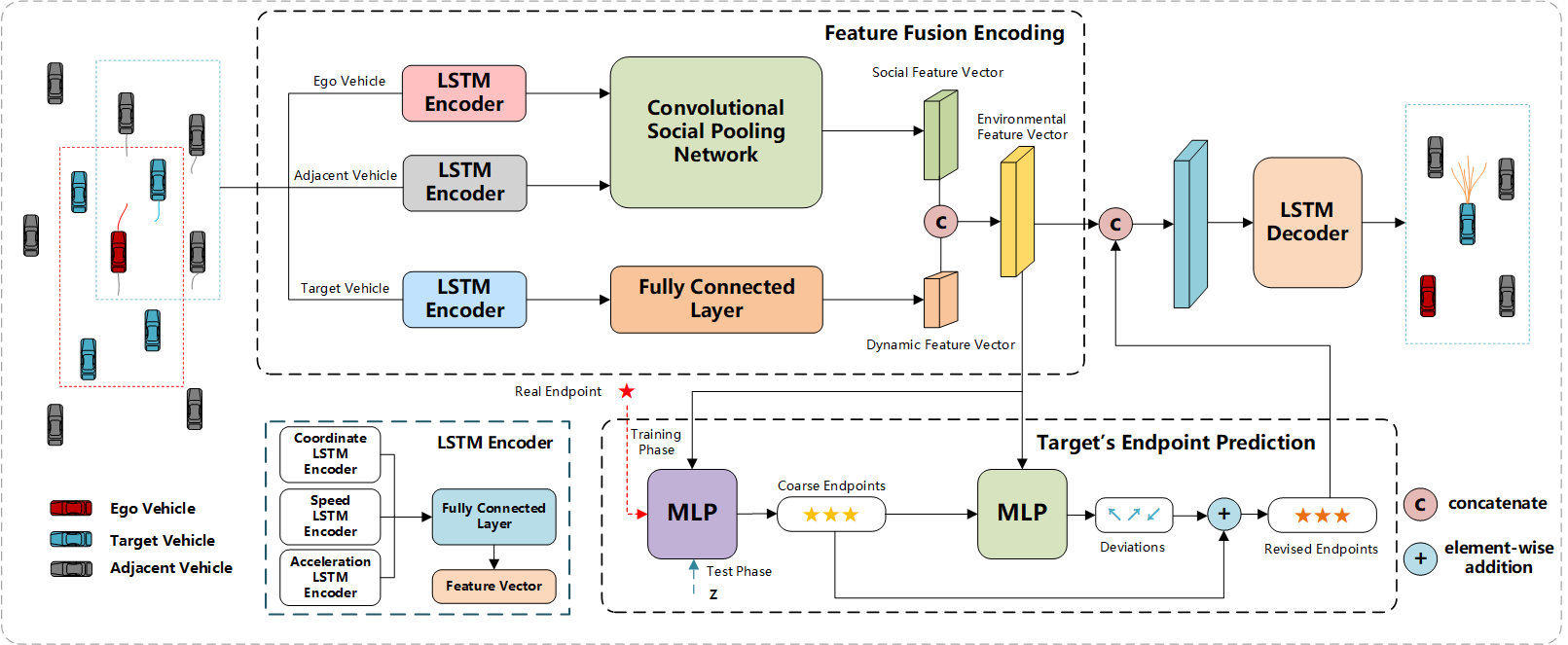}
 \caption{Overall architecture of EPN model.}
 \label{fig2}
\end{figure*}

This article introduces an Ego vehicle Planning-informed Network (EPN) for multimodal trajectory prediction, the model architecture is shown in Figure \ref{fig2}. EPN consists of three main modules: the feature fusion encoding, the target's endpoint prediction, and the LSTM decoder. The feature fusion encoding module separately encodes the information from the ego vehicle, target vehicle, and adjacent vehicles. The encoded features of the ego vehicle and adjacent vehicles are processed through a convolutional social pooling network to extract interaction information, generating social feature vectors. Simultaneously, the target vehicle's encoded features are transformed into dynamic feature vectors via a fully connected layer. These vectors are concatenated to form environmental feature vectors. The target's endpoint prediction module employs CVAE to predict potential target vehicle endpoints and refines them to improve accuracy. Finally, the LSTM decoder module combines the environmental and endpoint feature vectors to produce the target vehicle's complete multimodal predicted trajectory.

\subsection{Feature Fusion Encoding}
The feature fusion encoding module is designed to output an environmental feature vector that integrates the spatiotemporal information of the driving scene. This module encodes not only the historical trajectories and states of all vehicles in the scene but also the planned trajectory of the ego vehicle, fully considering its potential influence on the future trajectory of the target vehicle. Given LSTM’s effectiveness in handling temporal data, we utilize LSTM networks to encode vehicle information. Due to the limitations of the dataset and challenges in obtaining complete driving status data, we focus on using each vehicle's historical trajectory, speed, and acceleration as inputs. After processing this input information through an LSTM encoder, convolutional social pooling network, and fully connected layer, the final output is an environmental feature vector.

Since trajectory data, velocity, and acceleration have different dimensions, we apply distinct embedding layers with unique parameters for each, followed by separate LSTM encoders. Each vehicle’s historical trajectory and state data is initially passed through an embedding layer, then processed by an LSTM encoder, where the tensor from the final hidden layer represents the vehicle’s feature vector at the current time step. The specific computation process follows:

\begin{equation}
    h_p^i = LSTM(emb(x_t^i,y_t^i))
    \label{eq7}
\end{equation}
\begin{equation}
    h_v^i = LSTM(emb(v_t^i))
\end{equation}
\begin{equation}
    h_a^i = LSTM(emb(a_t^i))
    \label{eq9}
\end{equation}
in equations (\ref{eq7}) to (\ref{eq9}), $t = \left( {t - {T_h} + 1,t - {T_h} +2,...,t} \right)$, $(x_t^i,y_t^i)$, $v_t^i$, and $a_t^i$ denote the position coordinates, velocity, and acceleration of the $i$-th vehicle at time step $t$, respectively. Here, $h_p^t$, $h_v^t$, and $h_a^t$ represent the feature vectors associated with the final hidden layer of the LSTM encoder for position, velocity, and acceleration. Once these three feature vectors are obtained, a fully connected layer calculates the intrinsic relationships among the vehicle's position, velocity, and acceleration, integrating these features into a unified dimension. Additionally, a fully connected layer performs dimensionality reduction on the target vehicle’s feature vector, producing a dynamic feature vector that captures the target vehicle's temporal characteristics. The following equation details this computation:

\begin{equation}
    e_{nei}^t = FC(h_p^i,h_v^i,h_a^i), i \in {A_{nei}}
\end{equation}
\begin{equation}
    e_{tar}^t = FC(h_p^i,h_v^i,h_a^i), i \in {A_{tar}}
\end{equation}
\begin{equation}
    enc_{dyn} = FC(e_{tar}^t)
\end{equation}
here, $e_{nei}^t$ represents the final encoded feature vector of the vehicles surrounding the target vehicle at time $t$, which includes encoded information about the ego vehicle, while $e_{tar}^t$ is the final encoded feature vector of the target vehicle at time $t$. $enc_{dyn}$ denotes the dynamic feature vector of the target vehicle after dimensionality reduction. 

To encode the planned trajectory of the ego vehicle, we use embedding layers and LSTM networks to process its planning information:
\begin{equation}
    e_{plan}^t = LSTM(emb(x_t^i,y_t^i))
\end{equation}

To effectively model the interaction between the ego vehicle, adjacent vehicles, and the target vehicle while highlighting the influence of adjacent vehicles’ historical information and the ego vehicle's planned future trajectory on the target vehicle’s future path, we introduce a convolutional social pooling network to simulate inter-vehicle interactions within the scene. First, a mesh centered on the target vehicle is established using a masking mechanism, which collects the feature vectors of surrounding vehicles into a social tensor. The masking calculation formula is as follows:
\begin{equation}
    mas{k_{i,j}} = \left\{ \begin{array}{l}
1,{\rm{        }}if{\rm{  }}gri{d_{i,j}} = 1\\
0,{\rm{        }}if{\rm{  }}gri{d_{i,j}} = 0
\end{array} \right.
\end{equation}
here $gri{d_{i,j}}$ indicates whether an adjacent vehicle occupies position $\left( {i,j} \right)$ in the grid centered on the target vehicle. If $gri{d_{i,j}=1}$, it signifies the presence of an adjacent vehicle at that position, and $mas{k_{i,j}}$ is set to 1; otherwise, $mas{k_{i,j}}$ is set to 0. The resulting social tensor after masking adjacent vehicles is denoted as $t_{nei}$, while the social tensor for the planned trajectory of the ego vehicle is represented as $t_{plan}$.

The social tensor is subsequently input into the convolutional pooling network to extract interaction information between vehicles. For the social tensors representing adjacent vehicles and the ego vehicle planning, interaction information is extracted using the "convolution-pooling-convolution" operation, resulting in the intermediate tensors $en{c_{nei}}$ and $en{c_{plan}}$. These tensors are then concatenated through a pooling layer to generate the social feature vector, denoted as $en{c_{social}}$, which encapsulates the spatial feature information of the target vehicle. The formulas are as follows:

\begin{equation}
    en{c_{nei}} = ReLu(Conv2(MP1(ReLu(Conv1({t_{nei}})))))
\end{equation}
\begin{equation}
    en{c_{plan}} = ReLu(Conv2(MP1(ReLu(Conv1({t_{plan}})))))
\end{equation}
\begin{equation}
    en{c_{social}} = MP2(concat(en{c_{nei}},en{c_{plan}}))
\end{equation}

Finally, we combine the dynamic feature vector of the target vehicle with the social feature vector to create the environmental feature vector, denoted as $enc$. The formula is as follows:

\begin{equation}
    enc = concat(en{c_{social}},en{c_{dyn}})
\end{equation}

This environmental feature vector encompasses all relevant information derived from the raw data. Subsequent modules will utilize this environmental feature vector to predict the potential endpoint and future trajectory of the target vehicle.

\subsection{Target's Endpoint Prediction}
After obtaining the environmental feature vectors, we employ a CVAE to predict the potential endpoint positions that the target vehicle may reach. The CVAE can generate more appropriate sampling points by constraining the input data information to the variational autoencoder. To enhance the accuracy of the endpoints generated by the CVAE, we incorporate the true endpoint coordinates of the target vehicle into the model input.

In this module, the model extracts endpoint features using different processes at various stages. During the training phase, distinct multilayer perceptrons are employed as the endpoint encoder, latent variable encoder, and latent variable decoder. First, the endpoint encoder encodes the target vehicle's true endpoint. The resulting feature vector is concatenated with the environmental feature vector and passed to the latent variable encoder to produce a $latent$ variable. The $latent$ variable's mean $\mu $ and standard deviation $\sigma $ are calculated, and Gaussian noise $z$ is sampled from a normal distribution $N(\mu ,\sigma )$. Finally, $z$ is concatenated with the environmental feature vector and decoded by the latent variable decoder to generate the target vehicle's endpoint. The formulas for the training phase are provided in equations (\ref{eq19}) to (\ref{eq21}):

\begin{equation}
    en{d_{feature}} = {E_{end}}(G)
    \label{eq19}
\end{equation}
\begin{equation}
    latent = {E_{latent}}(concat(enc,en{d_{feature}})
\end{equation}
\begin{equation}
    \overline G  = {D_{latent}}(concat(enc,z))
    \label{eq21}
\end{equation}
where $G$ represents the true endpoint of the target vehicle, while ${E_{end}}$, ${E_{latent}}$, and ${D_{latent}}$ denote the endpoint encoder, latent variable encoder, and latent variable decoder, respectively. $en{d_{feature}}$ refers to the encoded endpoint feature vector, and $\overline G $ represents the predicted endpoint of the target vehicle. During the validation and testing phases, since the true endpoint of the target vehicle is unknown, we directly sample Gaussian noise $z$ from the latent space based on a normal distribution $N(0,{\sigma _T}I)$. The sampled noise $z$ is then concatenated with the environmental feature vector and decoded to generate potential endpoint predictions. Following the truncation trick in PECnet \cite{Mangalam2020}, we set the standard deviation ${\sigma _T=1.3}$. Additionally, by controlling the number of times the latent space is sampled, multiple possible endpoint coordinates for the target vehicle can be predicted.

Due to the significant error in directly predicting the target vehicle's endpoint, we employ an endpoint correction mechanism to improve the accuracy of the prediction. After the first-stage prediction results are generated, the correction mechanism is applied to refine the predicted endpoint. Specifically, we first use an endpoint encoder to re-encode the predicted endpoint $\overline G $ of the target vehicle. The resulting feature vector $end_{refine}$ is concatenated with the environmental feature vector $enc$ and input into an endpoint decoder ${D_{end}}$. This decoder generates a deviation value $d_{offset}$ that represents the offset between the predicted endpoint and the true endpoint. Finally, $d_{offset}$ is added to the predicted endpoint in the first stage to obtain the corrected endpoint $\widehat G$. The detailed calculation process is shown in equations (\ref{eq22}) to (\ref{eq24}):

\begin{equation}
    end_{refine} = {E_{end}}(\overline G )
    \label{eq22}
\end{equation}
\begin{equation}
    d_{offset} = {D_{end}}(concat(enc,end_{refine}))
\end{equation}
\begin{equation}
    \widehat G = \overline G  + d_{offset}
    \label{eq24}
\end{equation}

\subsection{LSTM Decoder}
In the LSTM decoder module, the endpoint encoder is used to encode the corrected endpoint $\widehat G$. The resulting encoded feature vector $\widehat {end}_{refine}$ is then concatenated with the environmental feature vector $enc$ and fed into the LSTM decoder. This process generates complete trajectories for each corrected endpoint $\widehat G$, resulting in multimodal trajectory prediction outcomes, the calculation process is shown in formulas (\ref{eq25}) and (\ref{eq26}):

\begin{equation}
    \widehat {end}_{refine} = {E_{end}}(\widehat G)
    \label{eq25}
\end{equation}
\begin{equation}
    \widehat Y = LSTM(concat(enc,\widehat {end}_{refine}))
    \label{eq26}
\end{equation}

Due to significant variations in the absolute coordinate values of vehicles across different scenarios, our aim is to mitigate the adverse effects of these absolute values on the performance of trajectory prediction models. To achieve this, we predict the relative displacement $\left( {\delta x_{t+\tau}^i,\delta y_{t+\tau}^i} \right)$ of the target vehicle over the next ${T_f}$ time steps, where $\tau  \in \left( {1,2,...,{T_f}} \right)$. By predicting the relative displacement, we can improve the accuracy of the predictions. Once the relative displacement is predicted, the absolute coordinates of the predicted trajectory can be calculated using the formulas (\ref{eq27}) and (\ref{eq28}):

\begin{equation}
    x_{t + T_f}^i = x_t^i + \sum\limits_{\tau  = 1}^{T_f} \delta  x_{t + \tau }^i
    \label{eq27}
\end{equation}
\begin{equation}
    y_{t + T_f}^i = y_t^i + \sum\limits_{\tau  = 1}^{T_f} \delta  y_{t + \tau }^i
    \label{eq28}
\end{equation}
here $x_t^i$ and $y_t^i$ represent the horizontal and vertical coordinates of the target vehicle at time step $t$, respectively.

\subsection{Loss Function}
We use the trajectory prediction error ${L_{pred}}$ and the CVAE error ${L_{cvae}}$ as loss functions. The trajectory prediction error ${L_{pred}}$ is calculated as the mean squared error (MSE) between the predicted trajectory $\widehat Y$ and the true trajectory $Y$, as well as between the corrected predicted endpoint $\widehat G$ and the true endpoint $G$. The CVAE used for endpoint prediction is trained using Kullback-Leibler (KL) divergence as its loss function. The specific loss functions are defined in equations (\ref{eq29}) and (\ref{eq30}):

\begin{equation}
    {L_{pred}} = {L_{mse}}(Y,\widehat Y) + {L_{mse}}(G,\widehat G)
    \label{eq29}
\end{equation}
\begin{equation}
    {L_{cvae}} = {D_{KL}}(N(\mu ,\sigma )||N(0,1))
    \label{eq30}
\end{equation}

\section{EXPERIMENTS}
\label{section5}
\subsection{Dataset and Data Preprocessing}
We trained and evaluated our model on two publicly available highway datasets: NGSIM \cite{Colyar2007}, \cite{Colyar2006} and HighD \cite{Krajewski2018highD}.

(1) NGIMS dataset: The NGSIM dataset was collected through a project initiated by the U.S. Federal Highway Administration. The dataset records vehicle position, speed, acceleration, and type at a sampling frequency of 10 Hz.

(2) HighD dataset: The HighD dataset was recorded using drones with cameras over six highways in Germany, offering an aerial perspective. It includes vehicle position, speed, and acceleration sampled at 25 Hz.

Given the different sampling frequencies and data characteristics of the two datasets, we performed data pre-processing to standardize them. For our experiments, we used a 3-second historical trajectory to predict a 5-second future trajectory, creating 8-second scene segments during pre-processing. To reduce data volume, the sampling frequency was downsampled to 5 Hz, yielding 40 time steps per scenario, with 15 steps for history (${T_h} = 15$) and 25 steps for prediction (${T_f} = 25$). The datasets were divided into training, validation, and testing sets in a 7:1:2 ratio.

\subsection{Experimental Setup and Evaluation}
The experimental setup for this study includes Python 3.7, PyTorch 1.7, and CUDA 11.7, with all experiments trained on an RTX 3080 GPU. The LSTM encoder is configured with a dimension of 64, while the LSTM decoder has a dimension of 128. The batch size is set to 64, and the model is trained for 15 epochs with a learning rate of 0.001. For comparison, the CL-LSTM and PiP methods used in this experiment categorize the driving intentions of the target vehicle into six classes, generating multimodal trajectory predictions based on these intentions. Accordingly, the number of predicted endpoints in the endpoint prediction module is set to $k = 6$ in this experiment.

Root Mean Square Error (RMSE) is the primary evaluation metric used on the NGSIM and HighD datasets, thus, we employ RMSE for model comparison and ablation studies. Additionally, we use Average Displacement Error (ADE) and Final Displacement Error (FDE) to further evaluate the model’s predictive accuracy. These metrics provide a comprehensive assessment of predictive performance.

\subsection{Model Comparison}
We compare and evaluate EPN against the following trajectory prediction models:

(1) S-LSTM \cite{Alahi2016Social}: Uses an LSTM encoder-decoder for vehicle trajectory prediction, with a fully connected social pooling layer to predict future trajectories.

(2) CS-LSTM \cite{Deo2018Conv}: Builds on S-LSTM with a convolutional social pooling layer to capture interactions and incorporates multimodal trajectory prediction based on horizontal and vertical driving intentions.

(3) S-GAN \cite{Gupta2018sgan}: Combines sequence prediction with GANs, generating multiple trajectory predictions and using the closest to the true future trajectory for evaluation.

(4) WSiP \cite{Wang2023Wsip}: Inspired by wave superposition, this model aggregates local and global vehicle interactions for dynamic social pooling.

(5) PiP \cite{Song2020pip}: Considers ego vehicle planning's impact on nearby vehicles’ trajectories using an LSTM encoder and a convolutional social pooling module.

(6) S-TF \cite{Liu2023stf}: Utilizes a sparse Transformer for multimodal prediction, incorporating trajectory, velocity, and acceleration information along with driving intentions (left offset, right offset, or straight).

(7) EPN: Our proposed model, which predicts multimodal trajectories based on endpoint correction and ego vehicle planning, selecting the trajectory closest to the true future trajectory for evaluation.

\begin{table}[]
\caption{RSME Comparison of Each Model on NGSIM Dataset}

\resizebox{\columnwidth}{!}{
\begin{tabular}{cccccccc}
\toprule
\multirow{2}{*}{\begin{tabular}[c]{@{}c@{}}Prediction \\ duration\end{tabular}} & \multicolumn{7}{c}{RSME} \\
\cmidrule(lr){2-8}
          &S-LSTM&CS-LSTM&S-GAN&WSiP&PiP&S-TF&EPN\\
          \midrule
\centering1s        & 0.60    & 0.58    & 0.57  & 0.56 & 0.55 & 0.99          & \textbf{0.38} \\
\centering2s        & 1.28   & 1.26    & 1.32  & 1.23 & 1.18 & 1.43          & \textbf{0.79} \\
\centering3s        & 2.09   & 2.07    & 2.22  & 2.05 & 1.94 & 1.7           & \textbf{1.12} \\
\centering4s        & 3.10    & 3.09    & 3.26  & 3.08 & 2.88 & 2.02        & \textbf{1.56} \\
\centering5s        & 4.37   & 4.37    & 4.40   & 4.34 & 4.04 & 3.33          & \textbf{2.31} \\
\bottomrule
\end{tabular}
}
\label{tab1}
\end{table}

\begin{table}[htbp]
  \centering
  \caption{RSME Comparison of Each Model on HighD Dataset}
  \resizebox{\columnwidth}{!}{
    \begin{tabular}{p{2.5em}ccccccc}
    \toprule
    \multirow{2}{*}{\begin{tabular}[c]{@{}c@{}}Prediction \\ duration\end{tabular}} & \multicolumn{7}{c}{RSME} \\
\cmidrule(lr){2-8}
          & S-LSTM & CS-LSTM & S-GAN & WSiP & PiP  & S-TF          & EPN           \\
          \midrule
    \centering1s    & 0.19  & 0.19  & 0.30   & 0.20   & 0.17  & 0.75  & \textbf{0.08} \\
    \centering2s    & 0.57  & 0.57  & 0.78  & 0.60   & 0.52  & 0.89  & \textbf{0.18} \\
    \centering3s    & 1.18  & 1.16  & 1.46  & 1.21  & 1.05  & 1.05  & \textbf{0.32} \\
    \centering4s    & 2.00     & 1.96  & 2.34  & 2.07  & 1.76  & 1.33  & \textbf{0.55} \\
    \centering5s    & 3.02  & 2.96  & 3.41  & 3.14  & 2.63  & 1.75  & \textbf{0.91} \\
    \bottomrule
    \end{tabular}
    }
  \label{tab2}
\end{table}

Tables \ref{tab1} and  \ref{tab2} present the RMSE values for different models on the NGSIM and HighD datasets, respectively. A lower RMSE indicates smaller prediction errors. Bold text highlights the method with the best performance among all compared models. As shown in Tables \ref{tab1} and \ref{tab2}, our proposed EPN model achieved the best predictive performance on the NGSIM and HighD datasets. 

When compared to models such as L-LSTM, CS-LSTM, S-GAN, and WSiP, which only consider historical trajectory information, the prediction accuracy of EPN is significantly improved. This demonstrates that incorporating the vehicle's driving status information and ego vehicle planning plays a crucial role in trajectory prediction. Furthermore, compared to the PiP model, which also accounts for the impact of ego vehicle planning on trajectory prediction, EPN exhibits a smaller prediction error, highlighting the benefits of our enhanced endpoint prediction module in improving model performance. In comparison with the S-TF model, which exhibits the smallest prediction error among the competing methods, EPN reduced the average RMSE by 34.9\% on the NGSIM dataset and 64.6\% on the HighD dataset. These results clearly demonstrate that our proposed EPN model outperforms the other models by a significant margin.

In addition to comparing the predictive performance of various models using statistical metrics, we also used Average Displacement Error (ADE) and Final Displacement Error (FDE) to further evaluate the performance of EPN and other multimodal trajectory prediction methods based on driving intentions. The comparison methods are as follows:

(1) CS-LSTM (M): For the multimodal trajectory prediction results based on different driving intentions output by the CS-LSTM method, we select the trajectory closest to the true future trajectory and use it to calculate the evaluation metrics.

(2) PiP (M): Similarly, for the multimodal trajectory prediction results based on different driving intentions from the PiP method, we select the trajectory closest to the true future trajectory to calculate the evaluation metrics.

(3) EPN: Our proposed multimodal trajectory prediction method, based on endpoint correction and ego vehicle planning, selects the trajectory closest to the true future trajectory from the predicted results to calculate the evaluation metrics.

\begin{table*}[htbp]
  \centering
  \caption{ADE/FDE Comparison of Each Model on NGSIM Dataset and HighD Dataset}
    \begin{tabular}{ccccccc}
    \toprule
    \multicolumn{1}{c}{\multirow{3}[5]{*}{Prediction {}duration}} & \multicolumn{6}{c}{ADE/FDE} \\
\cmidrule{2-7}          & \multicolumn{3}{c}{NGSIM} & \multicolumn{3}{c}{HighD} \\
\cmidrule{2-7}          & CS-LSTM(M) & PiP(M) & EPN   & CS-LSTM(M) & PiP(M) & EPN \\
    \midrule
    1s    & 0.19/0.35 & 0.19/0.34 & \textbf{0.12/0.24} & 0.09/0.16 & 0.08/0.14 & \textbf{0.04/0.07} \\
    2s    & 0.40/0.75 & 0.39/0.74 & \textbf{0.28/0.54} & 0.20/0.42 & 0.18/0.40 & \textbf{0.08/0.14} \\
    3s    & 0.58/1.13 & 0.58/1.11 & \textbf{0.41/0.76} & 0.34/0.79 & 0.37/0.81 & \textbf{0.12/0.24} \\
    4s    & 0.79/1.57 & 0.77/1.53 & \textbf{0.53/1.01} & 0.53/1.28 & 0.51/1.13 & \textbf{0.17/0.39} \\
    5s    & 1.03/2.41 & 1.00/2.21 & \textbf{0.69/1.58} & 0.75/1.88 & 0.69/1.64 & \textbf{0.24/0.63} \\
    \bottomrule
    \end{tabular}
  \label{tab3}
\end{table*}

Table \ref{tab3} compares the ADE and FDE of EPN with the multimodal trajectory prediction methods CS-LSTM (M) and PiP (M).

The experimental results show that when ego vehicle planning information is incorporated, the ADE and FDE values of PiP (M) are lower than those of CS-LSTM (M). Compared to PiP (M), EPN achieves an average reduction of 30.7\% in ADE and 30.4\% in FDE on the NGSIM dataset, and a more substantial reduction of 64.5\% in ADE and 64.3\% in FDE on the HighD dataset. These results demonstrate that our proposed multimodal trajectory prediction method, based on endpoint correction, outperforms the method based on predicted driving intentions. The performance improvement is especially significant on the HighD dataset, where the positioning data is more accurate, leading to a more pronounced reduction in prediction error.

\subsection{Ablation Experiment}
To evaluate the impact of each module in the EPN model on its predictive performance, we conducted ablation experiments with the following specific experimental settings:

(1) PCS-LSTM: Adds ego vehicle planning information to the CS-LSTM model.

(2) PCS-LSTM (V): Builds on PCS-LSTM by incorporating the vehicle's speed as a feature input.

(3) PCS-LSTM (V+A): Extends PCS-LSTM by encoding both the speed and acceleration of the vehicle as features, which are input into the model.

(4) PCS-LSTM (V+A+M): Uses the optimal trajectory from PCS-LSTM to calculate evaluation metrics.

(5) EPN (R): Adds an endpoint prediction module to the PCS-LSTM model, while removing the endpoint correction mechanism from the full model.

(6) EPN: The complete trajectory prediction framework proposed in this paper, based on endpoint correction and ego vehicle planning.

\begin{table*}[htbp]
  \centering
  \caption{RSME/ADE/FDE Comparison of Ablation Experiments on NGSIM Dataset}
    \begin{tabular}{>{\centering\arraybackslash}p{10em}ccccc} 
    \toprule
    \multirow{2}[4]{*}{Methods} & \multicolumn{5}{c}{RMSE/ADE/FDE} \\
\cmidrule{2-6}    \multicolumn{1}{c}{} & 1s    & 2s    & 3s    & 4s    & 5s \\
    \midrule
    PCS-LSTM & 0.55/0.20/0.37 & 1.19/0.43/0.87 & 1.95/0.69/1.44 & 2.90/0.98/2.13 & 4.07/1.30/2.98 \\
    PCS-LSTM(V) & 0.45/0.14/0.29 & 1.08/0.36/0.77 & 1.82/0.60/1.32 & 2.74/0.88/1.99 & 3.87/1.20/2.82 \\
    PCS-LSTM(V+A) & 0.42/0.13/0.26 & 1.03/0.33/0.73 & 1.74/0.57/1.27 & 2.62/0.84/1.92 & 3.72/1.15/2.72 \\
    PCS-LSTM(V+A+M) & 0.40/0.12/0.25 & 0.94/0.31/0.66 & 1.46/0.50/1.04 & 1.88/0.70/1.45 & 2.70/0.92/2.10 \\
    EPN(R) & 0.39/0.12/0.24  & 0.79/0.28/0.55 & 1.15/0.41/0.78 & 1.62/0.55/1.08 & 2.39/0.72/1.67 \\
    EPN   & \textbf{0.38/0.12/0.24} & \textbf{0.79/0.28/0.54} & \textbf{1.12/0.41/0.76 } & \textbf{1.56/0.53/1.01} & \textbf{2.31/0.69/1.58} \\
    \bottomrule
    \end{tabular}
  \label{tab4}
\end{table*}

Table \ref{tab4} presents the comparison results of RMSE, ADE, and FDE from the ablation experiments on the NGSIM dataset. Incorporating vehicle driving status information, such as speed and acceleration, improves trajectory prediction accuracy. Compared to the benchmark model, PCS-LSTM, the addition of speed as a feature in PCS-LSTM (V) results in average reductions of 6.6\%, 11.7\%, and 7.7\% in RMSE, ADE, and FDE, respectively. Further including both speed and acceleration in PCS-LSTM (V+A) leads to average reductions of 10.6\%, 16.1\%, and 11.4\% in these metrics. When comparing PCS-LSTM (V+A+M) to EPN (R), the latter shows an average reduction of 14.1\%, 18.4\%, and 21.5\% in RMSE, ADE, and FDE, respectively. This highlights the significant impact of the endpoint prediction module on improving the performance of multimodal trajectory prediction. Finally, by comparing EPN (R) with the full EPN model, we observe that the addition of the endpoint correction mechanism contributes to further improvements in the model’s predictive performance.

\begin{figure}[!t]
\centering
\includegraphics[height=5.6cm,width=8.5cm]{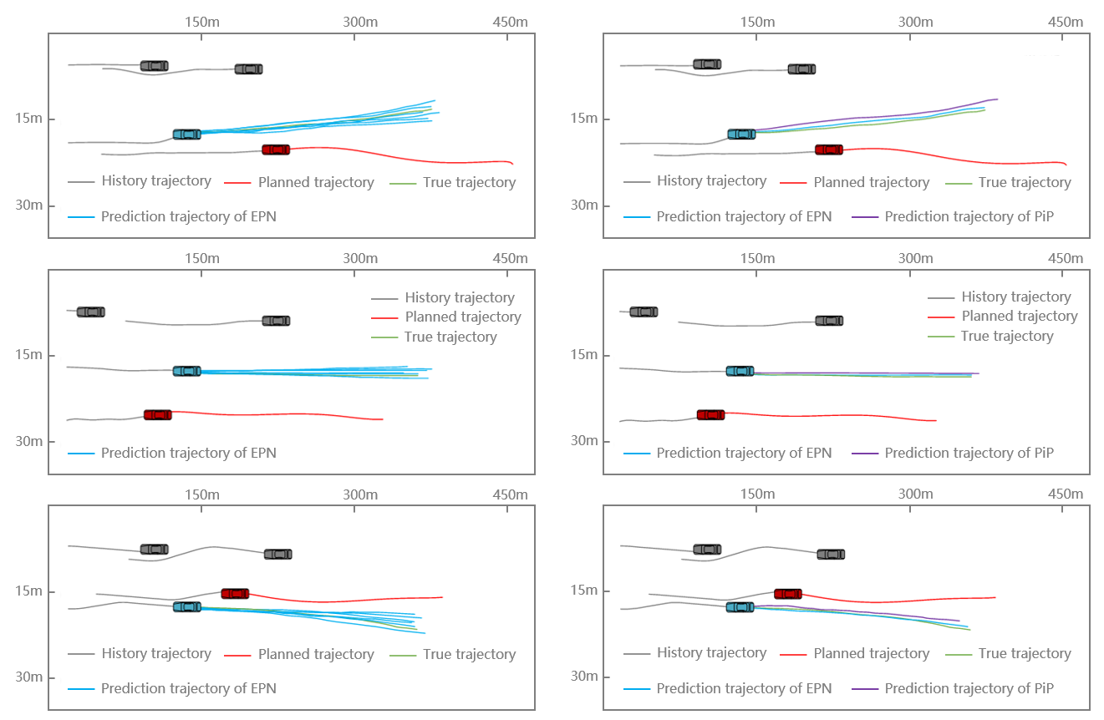}
 \caption{Visualization results in left-turn, straight-ahead and right-turn driving scenarios.}
 \label{fig3}
\end{figure}

\subsection{Visualization}
We visualize the trajectory data in different scenarios to qualitatively analyze the performance of the model. Figure \ref{fig3} presents the multimodal trajectory prediction results of the EPN in left turn, straight, and right turn driving scenarios for the target vehicle, along with a comparison of the optimal predictions between the EPN and PiP methods. Different colored curves are used to represent various trajectories.

From the visualizations, it is clear that the multimodal trajectory predictions of EPN closely align with the true future trajectory of the target vehicle, demonstrating strong prediction performance. While the predictions for straight driving are similar between EPN and PiP, in left and right turn scenarios, the EPN predictions more closely match the true trajectory than those of PiP. This is because, in turning scenarios, predicting the vehicle's possible endpoint directly, rather than relying on ambiguous driving intentions, enables a more accurate estimate of the vehicle's movement, resulting in predictions that better match the true trajectory.

\section{CONCLUSION}
\label{section6}
In this paper, we propose a trajectory prediction model based on ego vehicle planning. This model uses the historical trajectory, vehicle speed, acceleration, and planned future trajectory of the ego vehicle as inputs, and outputs a multimodal prediction of the target vehicle's future trajectory. By incorporating the planned trajectory of the ego vehicle, the interaction between the ego vehicle's plan and the predicted trajectory of the target vehicle can be simulated, thereby more realistically replicating the driving scenario and improving the accuracy of trajectory prediction for the target vehicle. We introduce a target's endpoint prediction module based on a CVAE to address the issues of intention ambiguity and large prediction errors typically found in trajectory prediction methods based on driving intentions. This module first predicts multiple potential endpoints for the target vehicle, and then refines the accuracy of these predictions through an endpoint correction mechanism. The module subsequently generates a complete multimodal trajectory based on the predicted endpoints. Experimental results demonstrate that our method achieves superior trajectory prediction accuracy compared to methods based solely on driving intentions.



\bibliographystyle{./IEEEtran} 
\bibliography{./IEEEabrv,./References}

\end{document}